\documentclass[letterpaper]{article} 
\usepackage{aaai25}  
\usepackage{times}  
\usepackage{helvet}  
\usepackage{courier}  
\usepackage[hyphens]{url}  
\usepackage{amsmath}
\usepackage{booktabs}
\usepackage{colortbl}
\usepackage{xr}
\usepackage{natbib}  
\usepackage{caption} 
\usepackage[table,xcdraw]{xcolor}
\usepackage{amsmath}
\usepackage{amsthm}
\usepackage{amsfonts}
\usepackage{dsfont}
\usepackage{caption}
\usepackage[ruled, linesnumbered]{algorithm2e}
\usepackage{tikz}
\usepackage{algpseudocode}
\usepackage{graphicx} 
\usepackage{subcaption}
\usepackage{array}
\usepackage{multirow}
\usepackage{diagbox}
\usepackage{multirow}

\title{SemiDFL: A Semi-Supervised Paradigm for Decentralized Federated Learning}
\author{
    Xinyang Liu\textsuperscript{\rm 1,\rm2}\thanks{Equal Contribution.},
    Pengchao Han\textsuperscript{\rm 3}\footnotemark[1],
    Xuan Li\textsuperscript{\rm 4},
    Bo Liu\textsuperscript{\rm 1}\thanks{Bo Liu is the corresponding author.}
}
\affiliations{
    \textsuperscript{\rm 1}Shenzhen Institute of Artificial Intelligence and Robotics for Society, China \\
    \textsuperscript{\rm 2}Department of Aeronautical and Aviation Engineering, The Hong Kong Polytechnic University, China\\
    \textsuperscript{\rm 3}School of Information Engineering, Guangdong University of Technology, China \\
    \textsuperscript{\rm 4}School of Information Science and Engineering, Southeast University, China \\

    codex.lxy@gmail.com, hanpengchao@gdut.edu.cn, xuanli2003@seu.edu.cn, liubo@cuhk.edu.cn
}

\usepackage{bibentry}

\begin{document}

\maketitle

\begin{abstract}
Decentralized federated learning (DFL) realizes cooperative model training among connected clients without relying on a central server, thereby mitigating communication bottlenecks and eliminating the single-point failure issue present in centralized federated learning (CFL). 
Most existing work on DFL focuses on supervised learning, assuming each client possesses sufficient labeled data for local training. However, in real-world applications, much of the data is unlabeled. We address this by considering a challenging yet practical semi-supervised learning (SSL) scenario in DFL, where clients may have varying data sources: some with few labeled samples, some with purely unlabeled data, and others with both.
In this work, we propose SemiDFL, the first semi-supervised DFL method that enhances DFL performance in SSL scenarios by establishing a consensus in both data and model spaces. Specifically, we utilize neighborhood information to improve the quality of pseudo-labeling, which is crucial for effectively leveraging unlabeled data. 
We then design a consensus-based diffusion model to generate synthesized data, which is used in combination with pseudo-labeled data to create mixed datasets. 
Additionally, we develop an adaptive aggregation method that leverages the model accuracy of synthesized data to further enhance SemiDFL performance.
Through extensive experimentation, we demonstrate the remarkable performance superiority of the proposed DFL-Semi method over existing CFL and DFL schemes in both IID and non-IID SSL scenarios.

\end{abstract}

\begin{links}
\link{Code}{https://github.com/ez4lionky/SemiDFL}
\end{links}

%

\section{Introduction}

Federated Learning (FL) \cite{mcmahan2017communication,kairouz2021advances} enables collaborative learning among distributed clients while keeping their data local and preventing ``data islands". 
However, traditional Centralized Federated Learning (CFL) relies on a central server to aggregate model parameters from all clients, which suffers from high communication bottleneck and single-point-failure problems \cite{8752023,8943129}.
Decentralized Federated Learning (DFL) \cite{beltran2023decentralized,sun2024byzantine} addresses this issue by enabling direct communication between clients and their connected neighbors for model aggregation. 
By removing the central server, DFL alleviates the communication bottleneck of traditional FL and enhances model scalability and robustness \cite{liu2021consensus}.

Existing CFL and DFL approaches are mostly designed for supervised learning, which relies heavily on large quantities of labeled data for model training. 
However, collecting and labeling extensive datasets in real-world applications is challenging due to time constraints, high costs, and the need for expert knowledge. Furthermore, clients often possess diverse data sources (e.g., labeled, unlabeled, or both) with heterogeneous distributions \cite{yang2022survey}. 
This raises this paper's key question: How can DFL be effective when clients have labeled and unlabeled data sources in highly non-iid scenarios?

Semi-supervised learning (SSL) \cite{yang2022survey} provides an effective approach to leverage unlabeled data for improved model performance. The main idea is to estimate pseudo-labels for unlabeled data, enhancing model training alongside data augmentation methods. 
Thus far, researchers have developed SSL methods for centralized federated learning scenarios \cite{liang2022rscfed,li2023class}, where a central server facilitates the sharing of supervised information among clients. However, applying these methods to DFL is challenging due to the lack of central coordination. To the best of our knowledge, no SSL method has been specifically designed for DFL with diverse data sources and non-IID data distributions.

Designing a practical and effective semi-supervised DFL method presents several challenges introduced by limited labels in non-IID settings.
First, accurately estimating pseudo-labels is crucial in SSL, but non-IID datasets make unbiased predictions challenging for a client's local model in DFL.
Second, relying only on labeled and pseudo-labeled data may be insufficient for model training particularly for clients with limited labeled data, while unbiased data generation in highly non-IID DFL scenarios is challenging.
In this paper, we propose a semi-supervised learning paradigm for DFL (SemiDFL) to deal with the above challenges, by establishing consensus in both model and data spaces.

Specifically, we propose a neighborhood pseudo-labeling method that introduces neighborhood classifiers for estimating pseudo-labels and uses neighborhood-qualified pseudo-label numbers to update the filtering threshold. 
This approach effectively improves the quality of pseudo-labeling and qualified pseudo-labeled data filtering, thereby enhancing classifier model training.
We then design a consensus-based diffusion model for each client to generate synthesized data with a similar data distribution for further data MixUp operation combined with labeled and pseudo-labeled data.
This forms a consensus data space among clients and therefore benefits the model training process and alleviates the non-IID issue.
Instead of using average aggregation in the global consensus process, we design an adaptive aggregation method based on each classifier's accuracy on synthesized data generated by its diffusion model. This approach enhances model aggregation efficiency, forming a more effective consensus model space for both classifier and diffusion, which alleviates data source divergence and improves overall model performance.

Our paper mainly makes the following contributions:

\begin{itemize}
	\item \textbf{SemiDFL Paradigm}: To our best knowledge, SemiDFL is the first semi-supervised DFL paradigm that is effective for clients with diverse data sources (labeled, unlabeled, and both) in highly non-IID settings. We innovatively propose consensus model and data spaces in  SemiDFL to address the challenges of limited labels and non-IID data in semi-supervised DFL scenarios.

    \item \textbf{Consensus data space}: We design neighborhood pseudo labeling to improve the pseudo labeling quality of each client by combining its neighborhood information, and utilize the consensus-based DFL rule to train a unified diffusion model to generate synthesized data for further data MixUp. This helps to build a consensus data space among all clients to boost the classifier training.

	\item \textbf{Consensus model space}: We design an adaptive consensus mechanism based on each classifier model's performance on generated data to dynamically aggregate neighborhood diffusion and classifier models, circumventing the need of an extra shared dataset for model performance evaluation. This helps establish a better consensus model space across diverse clients.
    
	\item \textbf{Extensive Experiments}: We comprehensively evaluate SemiDFL through extensive experiments on different datasets, models, SSL settings, and non-IID degrees. The results verified SemiDFL's superior performance against existing methods. 

\end{itemize}

\section{Preliminaries and Related Work}
\subsection{Decentralized Federated Learning}
Consider a DFL system with a set {$\mathcal{N}=\{1,2,\ldots,N\}$} of clients, where each client $i$ has a general local model ${\theta}_i$ and a private dataset $\mathcal{X}_i$ containing $\vert \mathcal{X}_i \vert$ data samples. The goal of DFL is to minimize the total loss across all $N$ clients:
\begin{equation}\label{eq:classifier}
	\min _{\Theta} \frac{1}{N} \sum_{i=1}^{N} l({\theta_i}; \mathcal{X}_i),
\end{equation}
where $\Theta = \{\theta_i, i \in \mathcal{N}\} $ is the set of all local $ \theta_i $, $l({\theta_i}; \mathcal{X}_i)$ represents the local loss function of client $i$. DFL typically runs over $ T $ rounds, with the following procedures executed sequentially in each round $ t $: 

\begin{itemize}
\item \emph{Local training:} Each client $i$ independently trains its local model on its dataset $ \mathcal{X}_i $. The update is given by:

\begin{equation}{\label{eq:local}}
	{{\theta}_i^{t+1/2}}={{\theta}_i}^t -\eta {\nabla {\theta}_{i}^t},
\end{equation}
where ${{\theta}_i^{t+1/2}}$ is the locally optimized model, $\eta$ is the learning rate, and ${\nabla \theta}_{i}^t$ is the gradient. Notably, it can be multiple local training iterations in each round $t$.

\item \emph{Global consensus:} 
Each client $i$ exchanges its local optimized model with its connected neighbors, and then aggregates the updates using a consensus mechanism: 
\begin{equation}{\label{eq:consensus1}}
	{{\theta}_i^{t+1}}=\sum_{j \in  \mathcal{G}_i}  w_{ij}{\theta}_j^{t+1/2},
\end{equation}
where $\mathcal{G}_i$ denotes the sub-graph centered on client $i$ (including client $i$ and its neighbors), $ w_{ij} $ is the aggregation weight {of client $j$ on client $i$}, $ {\theta}_i^{t+1} $ is the globally updated model for client $ i $ in round $ t+1 $.
\end{itemize}

\subsection{Semi-supervised Learning Objective in DFL}
We consider a practical SSL scenario with three types of clients based on their various data sources.

\begin{itemize}
	\item \emph{Labeled Client (L-client):} clients with only a few  labeled data samples.
	\item \emph{Unlabeled Client (U-client):} clients with only unlabeled data samples.
	\item \emph{Mixed Client (M-client):} clients with both a few labeled data samples and many unlabeled samples.
\end{itemize}
We consider a system consisting of  L-client set $ \mathcal N_L $, U-client set $ \mathcal N_U $, and M-client set $ \mathcal N_M $.
The aim of semi-supervised learning in DFL system is to jointly minimize the objectives of all clients as follows:
\begin{equation}
	\min _{\Theta}\!\frac{1}{N}\!(\!\sum_{i\in \mathcal{N}_L}\!l(\mathcal{L}_i;\theta_i)\!+\!\sum_{j\in \mathcal{N}_U}\!l(\mathcal{U}_j;\theta_j)\!+\!\sum_{k\in \mathcal{N}_M}\!l(\mathcal{M}_k;\theta_k)),
\end{equation}
where $ \mathcal{L}_i $, $ \mathcal{U}_i $ and $ \mathcal{M}_i $ represent the datasets of L-, U- and M-clients, respectively. 
For $ \mathcal{M}_i $, we further denote its labeled and unlabeled data as $ \mathcal{M}_i' $ and $ \mathcal{M}_i'' $, respectively.

\subsection{Related Works}
Semi-supervised learning (SSL) \cite{yang2022survey} aims to leverage unlabeled data to enhance machine learning model performance, especially when unlabeled data predominates. A classical SSL method is pseudo-labeling \cite{lee2013pseudo}, an entropy minimization approach \cite{grandvalet2004semi} that estimates a data sample's label based on model predictions. Pseudo-labeling is often combined with data augmentation techniques (e.g., MixUp \cite{zhang2017mixup}, MixMatch \cite{berthelot2019mixmatch}, FixMatch \cite{sohn2020fixmatch}) to boost SSL performance. The main idea behind data augmentation is to expand the training dataset by interpolating data samples. Recent work \cite{you2024diffusion} proposes using a diffusion model to generate synthesized data, further improving data augmentation in SSL.

Existing federated learning approaches primarily focus on supervised learning with fully labeled datasets, which may be impractical in many real-world scenarios, as noted in \cite{jin2020towards}. One research direction assumes a labeled server with unlabeled clients \cite{diao2022semifl,diao2021semifl,albaseer2020exploiting}. In \cite{jeong2021federated}, model parameters are divided between servers with labeled data and clients with unlabeled data. The study \cite{zhang2021improving} suggests training and aggregating a server model with labeled data to guide parallel client models with unlabeled data.

Another research direction assumes that some clients have labeled data. The work \cite{fan2022private} proposes using a shared GAN model to generate synthesized data, creating a unified data space to enhance federated SSL performance. RSCFed \cite{liang2022rscfed} tackles the challenges of federated semi-supervised learning in non-iid settings by employing random sub-sampling and distance-reweighted model aggregation. CBAFed \cite{li2023class} uses a fixed pseudo-labeling strategy to prevent catastrophic forgetting and designs class-balanced adaptive thresholds based on local data distributions to address SSL challenges in the federated learning setting.

However, without a central coordinator server, it is challenging to apply SSL methods for FL to DFL scenarios with varying data sources and non-IID data distribution. 
To our knowledge, there is currently no SSL method specifically tailored for DFL that addresses the challenges of diverse data sources and non-IID data distributions.

\section{Methodology}
\begin{figure*}[t]
\centering
\includegraphics[width=1.0\textwidth]{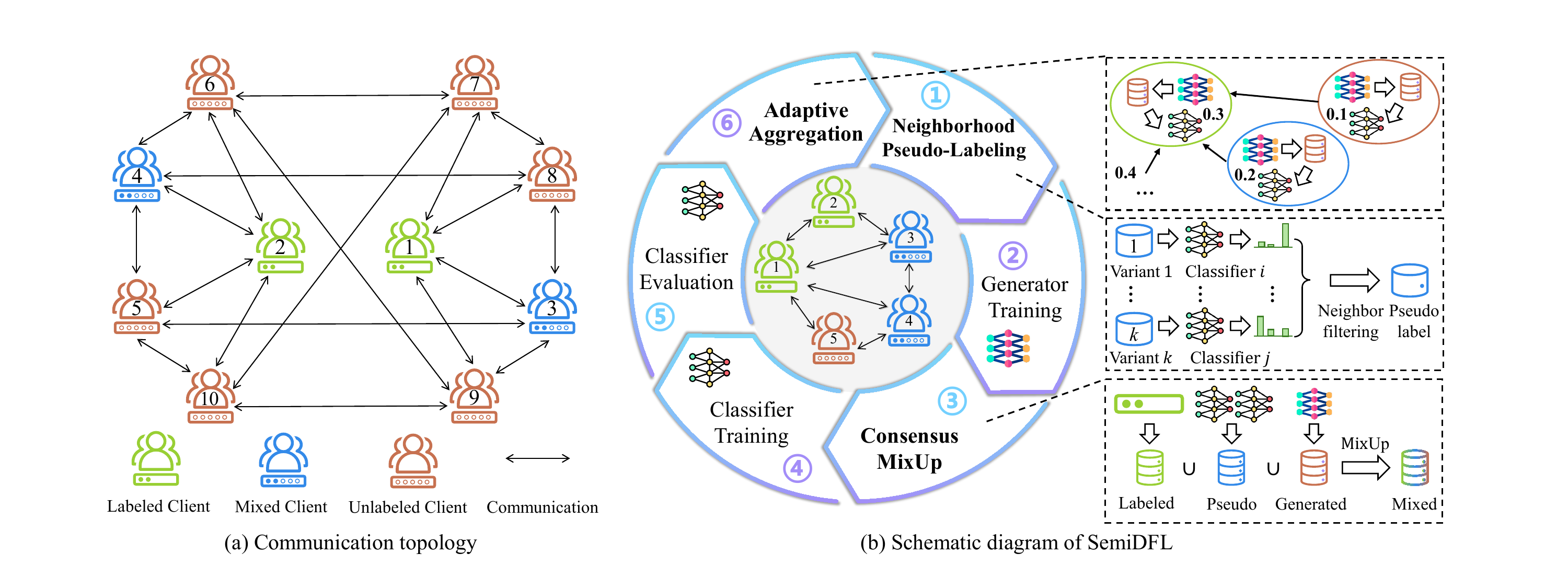}
\caption{Framework of SemiDFL. (a) is an example of a decentralized communication topology employed in our experiments; (b) illustrates the overall process of the proposed SemiDFL, which consists of six main steps as indexed. Among these steps, steps 2, 4, and 5 are general ideas, while steps 1, 3 and 6 are the main contributions of our work. The detailed working flows are illustrated on the right of (b).}
\label{overview}
\end{figure*}
\subsection{Overview of SemiDFL}
SemiDFL aims to establish consensus in both model and data spaces among connected clients, addressing the challenge of limited labels in semi-supervised DFL without data sharing. It achieves consensus model space through global consensus on classifiers and diffusion models (presented by $\phi_i$ and $\psi_i$ respectively, $i\in\mathcal{N}$) during the DFL training process.
All consensus-based diffusion models generate synthesized data following a similar distribution, which is then mixed with labeled and pseudo-labeled data to form a consensus data space.
The framework of SemiDFL is illustrated in Figure \ref{overview}, with key components described below:

\begin{itemize}
	\item \textbf{Neighborhood Pseudo-Labeling}: After training local classifier model $\phi_{i}, i\in\mathcal{N}$,  client $ i $ first estimates pseudo-labels of its unlabeled samples (if it has) using local model $\phi_{i}$ and then filters unlabeled samples with qualified pseudo-label. 
	Notably, neighborhood classifiers are introduced to enhance the pseudo-labeling process, and neighborhood-qualified pseudo-label numbers are used to adaptively update the filtering threshold.
	
	\item \textbf{Consensus MixUp}: Following pseudo-labeling, each client $ i $ trains a diffusion model $ \psi_{i} $ using labeled and/or pseudo-labeled samples. 
    Notably, all diffusion models $ \psi_{i}, i\in \mathcal{N} $ are aggregated based on a consensus mechanism, leading to a unified diffusion model. This makes synthesized data $\mathcal{D}_i, i \in \mathcal{N}$ generated from diffusion models $\psi_i, i \in \mathcal{N}$ follows a similar distribution (consensus data space) for subsequent MixUp.
	
	\item \textbf{Adaptive Aggregation}: Instead of using constant weights for classifier and diffusion models aggregation to form consensus model space, each client $i$ adaptively aggregates its neighborhood models {$w_{ij}$} based on classifier performance.	
	Each client $ i $ samples an extra small dataset from $\mathcal{D}_i, i \in \mathcal{N}$ (share a similar distribution) to evaluate its classifier. 
    This not only ensures that all local models are evaluated on a shared dataset but also helps to circumvent data island problems in DFL.
\end{itemize}

\subsection{Neighborhood Pseudo-Labeling}
\subsubsection{Pseudo-Labeling}

For the $n$-th indexed unlabeled data sample $u_i^n \in \mathcal{U}_i$ or $u_i^n \in \mathcal{M}_i''$ of client $i$, we estimate its pseudo-label using the client’s classifier model $\phi_i$, combined with label-invariant data augmentation and a label sharpening method \cite{berthelot2019mixmatch}. 
We first compute the label prediction $\Bar{p}_{n}^i \in \mathbb{R}^{1 \times C}$ ($C$ is the number of categories) for sample $u_i^n$ as follows:

\begin{equation}
	\Bar{p}^{n}_i = \frac{1}{K} \sum_{k=1}^K F(u^{n}_{i,k}; \phi_i),
\end{equation}
where $K$ is the number of variants for data augmentation, such as image rotation and cropping, $ F(u^{n}_{i,k}; \phi_i) $ is the classifier with $u^{n}_{i,k}$  being the $k$-th augmented variant of sample $u_i^n$. 

We then estimate the pseudo-label {of $u_i^n$} using label sharpening. The probability $\hat{p}_{i}^{n, c}$ that $u_i^n$ is predicted as class $c$ is given by:

\begin{equation}\label{eq:prediction}
	\hat{p}_{i}^{n,c} = \frac{(\Bar{p}_{i}^{n,c})^Z}{\sum_{c'=1}^C (\Bar{p}_{i}^{n,c'})^Z},
\end{equation}
where $\Bar{p}_{i}^{n, c}$ is the probability of class $c$ in $\Bar{p}_{i}^{n}$, and $Z > 1$ amplifies the dominating classes.

\subsubsection{Neighborhood Pseudo-Labeling (NPL)}

In the challenging SemiDFL scenario with non-IID data, traditional pseudo-labeling methods suffer from high divergence among clients and noisy pseudo-labels, making them inadequate for providing high-quality supervision independently. We propose leveraging neighborhood information to address these challenges.

On the one hand, we introduce neighborhood classifiers to predict the label of a sample $u_i^n$'s variant in client $i$. This design combines client $ i $ and its neighborhood classifiers' label predictions, thereby enhancing the robustness of pseudo-label prediction and reducing instability in highly non-IID settings.
Specifically, we modify the label prediction as:

\begin{equation}
	{\Bar{p}^{n}_i} = \frac{1}{K} (F(u^{n}_{i,1}; \phi_i) + \sum_{k=2}^K F(u^{n}_{i,k}; \phi_k)),
\end{equation}
where $\phi_k$ is a randomly selected model from the {\color{black}neighborhood set $\mathop{\mathcal{G}_i }$ of client $i$, i.e.,  $\left\{\phi_j, j \in \mathop{\mathcal{G}_i }\right\}$,} for each data variant $u^{n}_{i,k}$.

On the other hand, we design a neighborhood adaptive class-wise threshold to filter noisy pseudo-label predictions, inspired by the dynamic threshold determination in \cite{li2023class}. First, we calculate the number of samples with pseudo-label prediction probability above a manually defined threshold $\tau$, given by:

\begin{equation}
\sigma^{t,c}_i\!=\!\sum_{n=1}^{|\mathcal{U}i|}\!\mathds{1}\!\left(\!\max_c(\hat{p}_{i}^{n,c})\!>\!\tau\!\right)\!\cdot\!\mathds{1}\left(\arg \max_c (\hat{p}_{i}^{n,c}) = c \right),
\end{equation}
where $\sigma^{t,c}_i$ is the number of samples of all the unlabeled data in current client $i$ with pseudo-label prediction probability above the threshold in class $c$ and $\mathds{1}(\cdot)$ is an indicator, taking $1$ if the condition in the parentheses is true, and $0$ otherwise.

Then, the threshold is adaptively updated by:

\begin{equation}\label{eq:adaptive threshold}
	\tau^{t,c}_{i} = \frac{\sigma^{t,c}_i}{\max \limits_{i} ({\max \limits_{c} (\sigma^{t,c}_i), i \in \mathcal{G}_i})} \tau,
\end{equation}
where $ \tau^{t,c}_i $ is the neighborhood adaptive threshold.
This helps balance the number of pseudo-label samples, reducing the negative impact of non-IID data.

\subsection{Consensus MixUp}
\subsubsection{Local MixUp (L-MixUp)}
After filtering unlabeled samples with qualified pseudo-labels for client $i$, we generate mixed samples using MixUp based on the union set of labeled dataset $\mathcal{L}_i$ and pseudo-labeled samples $\mathcal{P}_i$ for local classifier model training. Suppose $(x_m, y_m)$ and $(x_n, y_n)$ are two $m$- and $n$-th indexed data samples in the union set $ \mathcal{L}_i \cup \mathcal{P}_i$. The MixUp operation produces:

\begin{equation}
	\begin{aligned}
		& x' = \lambda x_m + (1-\lambda)x_m,\\
		& y' = \lambda y_m + (1-\lambda)y_n,
	\end{aligned}
\end{equation}
where $\lambda$ is a hyper-parameter that determines the mixing ratio, $(x', y')$ is the mixed data sample.

\subsubsection{Consensus MixUp (C-MixUp)}
L-MixUp leads to suboptimal performance in the DFL scenarios with few and non-IID labeled data because data sharing is not allowed among agents.
For this issue, we propose the consensus MixUp (C-MixUp) which facilitates data imputation across all agents without raw data sharing. 

C-MixUp employs a generative learning manner to form a consensus data space, enabling the generation and mixing of data samples without data sharing.
For each client $ i $, we adopt a diffusion model to generate synthesized data \cite{sohl2015deep,ho2020denoising}. 
In the training phase, we first obtain the noisy latent $X_h$ by progressively adding noise to input data $X_0$, where $h$ denotes a randomly sampled diffusion timestep. Then, We train a diffusion model $D(\cdot;\psi_i)$ to predict the noise $\epsilon$ added on $X_h$ by minimizing the following objective $l^\prime(\cdot;\psi_i)$:
\begin{equation}\label{eq:diffusion}
	l^\prime(\cdot;\psi_i)=\mathbb{E}_{h,X_0,\epsilon \sim \mathcal{N}(0, \mathrm{I})}[\Vert D(X_h,c, h;\psi_i) - \epsilon \Vert^2_2]
\end{equation}
where $ \psi_i $ is parameter of the diffusion model, $ x $ is a data sample with $ c $ being its class. Notably, the $X_0$ could come from the labeled or unlabeled dataset, and $c$ maybe its real class or generated pseudo-label.

In the sampling phase, pseudo-images are generated via deterministic sampling \cite{song2020denoising, lu2022dpm} combined with the classifier-free method \cite{ho2022classifier} from the trained diffusion model.
We denote this generated dataset by diffusion model as $\mathcal{D}_i$. 
Notably, the local diffusion model $\psi_i$ is globally updated through a consensus mechanism during the DFL training process, as shown in (\ref{eq:consensus1}). This ensures that all local $\psi_i$ converge to a unified model, thus all generated datasets $\mathcal{D}_i, i \in \mathcal{N}$ follow a similar data distribution.

The generated data from $\mathcal{D}_i$ is then combined with $\mathcal{X}_i$ and $\mathcal{P}_i$ to produce more mixed samples for enhancing local training. Specifically, we create a new union set $\mathcal{L}_i \cup \mathcal{P}_i \cup \mathcal{D}_i$ to improve the MixUp process. This forms a consensus data space.

\subsubsection{Adaptive Aggregation} 
In the vanilla DFL setting, each agent's aggregation weight is fixed, based on the communication topology and consensus strategy. This unchanging influence on neighbors is inefficient in the DFL-Semi scenario. Ideally, agents with better performance should exert more influence, i.e., have larger aggregation weights. 
However, assuming a shared test dataset to evaluate each client's performance is impractical due to privacy constraints.

Nevertheless, since all datasets $\mathcal{D}_i, i \in \mathcal{G}_i$ share a similar distribution, making them ideal for performance evaluation.
To reduce computing cost, each client randomly samples an tiny validation dataset ($100$ samples in our experiments)  $\mathcal{\hat D}_i^t $ from $ \mathcal{D}_i $ to evaluate its performance for determining the aggregation weights of neighbors at the $t$-th round.

We use $ a_i $ to denote the accuracy of model $\phi_{i}$ on the validation dataset $ \mathcal{\hat D}_i^t $. The adaptive weight is then obtained by

\begin{equation}\label{eq:adaptive weight}
	w_{ij}  =  \frac{\exp(a_j - \bar a_{_i})}{\sum_{j \in \mathcal{G}_i}\exp(a_j - \bar a_{i})},
\end{equation}
where $ \bar a_{i} = \frac{1}{\vert \mathcal{G}_i\vert } \sum_{j \in \mathcal{G}_i}  a_j$ is the averaged accuracy in the sub-graph $ \mathcal{G}_i $ with $ \vert \mathcal{G}_i\vert $ being the number of clients in the sub-graph.

Based on the above three main components, we proposed SemiDFL as detailed in \textbf{Algorithm} \ref{alg:semidfl} \footnote{ The variable $Y_i$ and $Z_i$ are to denote client $i$'s label data and unlabeled data, respectively. The symbols $ \mathcal{M}_i' $ and $ \mathcal{M}_i'' $ represent the labeled and unlabeled data of M-client $ i $, respectively.}.

\begin{algorithm}[t]
	\caption{SemiDFL}\label{alg:semidfl}
	\textbf {Inputs:} Client number $N$, L-client set $\mathcal{N}_L$, U-client set $\mathcal{N}_U$, M-client set $\mathcal{N}_M$, communication topology $\mathcal{G}$, global training rounds $T$, diffusion warm-up round $ R $.
	
	\textbf {Outputs:} Optimal classifier $ {\phi}_i $ for client $i \in \mathcal{N}$.
	
	\textbf {Initialization:} Local classifier model $ {\phi}_i^0 $ and local diffusion model $ {\psi}_i^0 $ for  client $i \in \mathcal{N}$. \\

	\For{ round $ t \in \{1,2,\ldots,T\} $}
	{
		
		\For {client $ i \in \mathcal{N} $ \textbf {in parallel}}
		{            
			\textbf{if} $ i \in \mathcal{N}_L$ \textbf{then} $ Y_i = \mathcal{L}_i  $; $ Z_i = \emptyset $. \\
			\ \vline \ \ \textbf{else if} $ i \in \mathcal{N}_U$  \textbf{then} $ Y_i = \emptyset $; $ Z_i = \mathcal{U}_i $. \\
			\ \vline \ \ \textbf{else if} $ i \in \mathcal{N}_M$  \textbf{then} $ Y_i = \mathcal{M}_i' $; $ Z_i = \mathcal{M}_i''. $ \\
			\textbf{end}\\
			
			$\mathcal{P}_i^t \gets $ NPL(${Z}_i$,$ \{\phi_j^t, j \in \mathop{\mathcal{G}_i }\} $) by (\ref{eq:prediction})-(\ref{eq:adaptive threshold}).\\
			
			\footnotesize {${\psi}_{i}^{t+\frac{1}{2}} \gets$ Train diffusion on $Y_i \cup \mathcal{P}_i^t $ by (\ref{eq:diffusion}), (\ref{eq:local})}.\\
			
			\If{$ t \geq R$}{
			$\mathcal{D}_i^t \gets$ Generate data from diffusion $ {\psi}_{i}^{t+\frac{1}{2}}$. \\
			}
			$ {S_i^t} $ $ \gets $ C-MixUp($ Y_i \cup \mathcal{P}_i^t  \cup \mathcal{D}_i^t, \lambda $).\\
			
			${\phi}_{i}^{t+\frac{1}{2}} \gets$ Train classifier on $S_i^t$ by (\ref{eq:classifier}), (\ref{eq:local}).
			
			$\hat{\mathcal{D}}_i^t \gets$ Small validation dataset from $ \mathcal{D}_i^t $.\\
			$ a_i \gets$ Model evaluation of $ {\phi}_{i}^{t+1/2} $ on $ \hat{\mathcal{D}}_i^t$.
		}

		\For {client $ i \in \mathcal{N} $ \textbf {in parallel}}
		{
			$ w_{ij}^t, j\in \mathcal{G}_i \gets $ Compute  weight by (\ref{eq:adaptive weight}).\\
			$ {\phi}_{i}^{t+1}\gets$ Consensus update ${\phi}_{i}^{t+1/2}$ by (\ref{eq:consensus1}), \\
			${\psi}_{i}^{t+1} \gets$ Consensus update ${\psi}_{i}^{t+1/2}$ by (\ref{eq:consensus1}).
		}
	}
\end{algorithm}

\section{Experiments}
\begin{table*}[ht!]
\centering
\footnotesize
\caption{Averaged accuracy $\pm$ Standard deviation of all clients on MNIST, Fashion-MNIST, and CIFAR-10 datasets. (DFL-UB assumes all data are labeled, so it remains unchanged with varying $r$. The symbol ``-----"  denotes a non-convergent result.)}
\label{main_table}
\begin{tabular}{cccccccc}

\toprule
\multirow{2}{*}{\diagbox[dir=SE]{\tiny Method}{\tiny Setting}} & \multirow{2}{0.1\textwidth-2\tabcolsep - 1.25\arrayrulewidth}{\centering non-IID Degree} & \multicolumn{2}{c}{MNIST} & \multicolumn{2}{c}{Fashion-MNIST} & \multicolumn{2}{c}{CIFAR-10} \\ 
\cmidrule(l){3-4} \cmidrule(l){5-6} \cmidrule(l){7-8}
& & $r=0.5\%$ &  $r=0.1\%$ & $r=0.5\%$ & $r=0.1\%$ & $r=5\%$ & $r=1\%$ \\ \toprule
\multirow{2}{*}{DFL-UB} 
& $\alpha=100$ & \multicolumn{2}{c}{$95.76\pm0.10$}& \multicolumn{2}{c}{$86.12\pm0.14$} & \multicolumn{2}{c}{$86.11\pm0.08$}\\ 
& $\alpha=0.1$ & \multicolumn{2}{c}{$94.27\pm0.76$}& \multicolumn{2}{c}{$73.64\pm3.58$} & \multicolumn{2}{c}{$58.75\pm4.12$}\\ 
\midrule
\multirow{2}{*}{DFL-LB} 
& $\alpha=100$ & $89.24\pm0.06$ & $68.98\pm0.02$ & $65.16\pm1.55$ & $48.53\pm1.76$ & $55.96\pm0.11$ & $41.30\pm0.03$ \\ 
& $\alpha=0.1$ & $82.36\pm1.18$ & $66.49\pm0.05$ & $49.08\pm5.46$ & ----- & $16.96\pm0.27$ & ----- \\
\midrule
\multirow{2}{*}{MixMatch} 
& $\alpha=100$ & $86.89\pm0.69$ & $61.13\pm18.27$ & $65.97\pm1.65$ & $35.82\pm1.75$ & $65.54\pm0.23$ & $48.16\pm0.16$\\ 
& $\alpha=0.1$ & $63.69\pm14.62$ & $39.40\pm10.90$ & $54.52\pm3.14$ & $42.46\pm4.28$ & $44.49\pm1.15$ & $30.53\pm1.37$\\
\midrule
\multirow{2}{*}{FlexMatch} 
& $\alpha=100$ & $91.85\pm0.23$ & $71.84\pm1.03$ & $69.33\pm1.12$ & $52.40\pm0.26$ & $56.90\pm0.18$ & $37.91\pm0.20$\\ 
& $\alpha=0.1$ & $81.38\pm2.58$ & $55.96\pm2.31$ & $57.63\pm2.37$ & $40.03\pm4.05$ & $46.09\pm0.50$ & -----\\
\midrule
\multirow{2}{*}{CBAFed} 
& $\alpha=100$ & $69.93\pm1.61$ & $40.67\pm0.44$ & $55.54\pm1.16$ & $27.99\pm0.22$ & $42.29\pm0.91$ & $26.18\pm0.23$\\ 
& $\alpha=0.1$ & $70.65\pm4.64$ & $37.79\pm2.48$ & $47.39\pm4.40$ & ----- & $31.93\pm1.13$ & $21.50\pm0.89$\\
\midrule
\rowcolor{gray!20} 
& $\alpha=100$ & $\mathbf{94.46\pm0.06}$ & $\mathbf{90.88\pm0.07}$ & $\mathbf{75.36\pm0.27}$ & $\mathbf{60.27\pm0.16}$ & $\mathbf{69.40\pm0.20}$ & $\mathbf{49.79\pm0.18}$ \\
\rowcolor{gray!20} \multirow{-2}{*}{\textbf{SemiDFL}} 
& $\alpha=0.1$ & $\mathbf{88.69\pm0.09}$ & $\mathbf{70.66\pm0.50}$ & $\mathbf{71.49\pm0.14}$ & $\mathbf{49.40\pm0.71}$ & $\mathbf{49.63\pm0.61}$ & $\mathbf{40.28\pm0.89}$ \\ 
\bottomrule
\end{tabular}
\end{table*}

\subsection{Experimental Setup}
We use the decentralized communication topology in Figure~\ref{overview}(a) as an example to evaluate SemiDFL on different datasets, various labeled data ratios, and non-IID degrees.
More detailed experimental settings and parameters can be found in the supplementary material.

\subsubsection{Datasets and Models} 
We evaluate SemiDFL on MNIST~\cite{lecun1998mnist} and Fashion-MNIST~\cite{xiao2017fashion} using Convolutional Neural Network (CNN), and on CIFAR-10~\cite{krizhevsky2010cifar} using ResNet-18. 

\subsubsection{Hyper-Parameters}
We adopt the same method in \cite{hsu2019measuring,lin2020ensemble} 
to simulate different non-IID data distribution degrees. Specifically, the non-IID degree is captured by a sample allocation probability $\alpha$, with smaller $\alpha$ indicating a higher non-IID degree. 
The percentage of total labeled data in the union of all clients' data is denoted by the labeled data ratio $r$.
We train models for $500$ global rounds on all datasets. 
In each global training round, each client performs $E$ ($25$ for MNIST, $50$ for Fashion-MNIST and CIFAR-$10$) iterations of local training via mini-batch SGD with a batch size of $B = 10$. 
Other hyper-parameters during local model training are inherited from the default settings of Adam~\cite{kingma2014adam} and for all datasets, we use a learning rate of 0.05. 
All experiments are conducted using PyTorch $2.0$ on a machine with $2$ RTX $4090$ GPUs.

\subsubsection{Baseline methods}
We use the following SSL methods as baselines by adapting them to fit the DFL scenario.

\begin{itemize}
    \item \textbf{DFL upper bound (DFL-UB)}: This represents the DFL upper bound, where the model is trained assuming all data are labeled.
    \item \textbf{DFL Lower Bound (DFL-LB)}: This represents the DFL lower bound, where only the partially labeled data is available for model training.
    \item \textbf{MixMatch\cite{berthelot2019mixmatch}}: MixMatch assigns low-entropy labels to the augmented unlabeled data samples and blends them with labeled data using MixUp.
    \item \textbf{FlexMatch~\cite{zhang2021flexmatch}}: FlexMatch improves Fixmatch~\cite{sohn2020fixmatch} by flexibly adjusting class-specific thresholds at each round to select informative pseudo-labeled data.
    \item \textbf{CBAFed~\cite{li2023class}}: CBAFed addresses catastrophic forgetting with fixed pseudo-labeling, uses class-balanced adaptive thresholds for training balance, and employs residual weight connections for optimized model aggregation.

\end{itemize}

\subsection{Comparison Study}

Table \ref{main_table} presents the quantitative results of our SemiDFL alongside the state-of-the-art baselines on three datasets. Our proposed SemiDFL consistently outperforms all baselines (except DFL-UB) across different non-IID degrees $\alpha$ and various labeled data ratios $r$.
Notably, SemiDFL achieves higher averaged accuracy and a smaller standard deviation of all clients compared with other SSL methods, benefiting from its established consensus model and data space.
This demonstrates the effectiveness and superiority of SemiDFL in handling diverse semi-supervised DFL tasks with varying data settings.

\subsubsection{Robustness on various non-IID degrees}
\begin{figure}[t]
\centering
\begin{subfigure}{0.48\linewidth}
	\centering
	\includegraphics[width=1\textwidth]{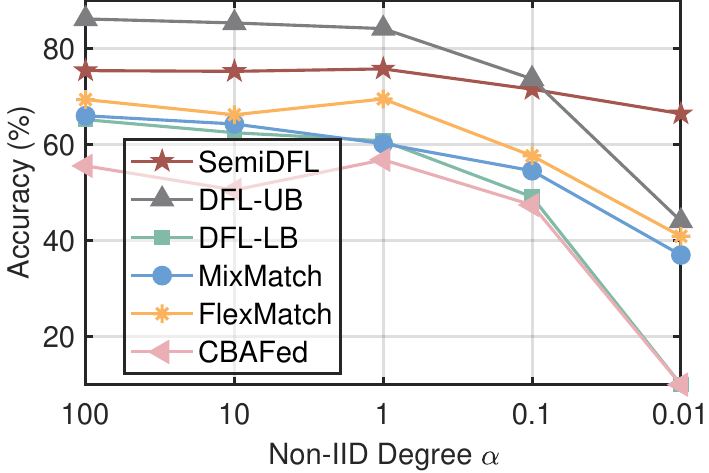}
	\caption{Fashion-MNIST ($r\!=\!0.5\%$).}
\end{subfigure}
\begin{subfigure}{0.48\linewidth}
	\centering
	\includegraphics[width=1\textwidth]{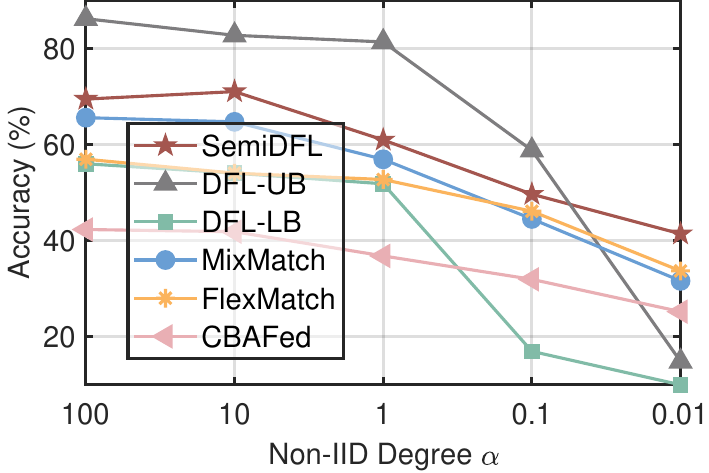}
	\caption{CIFAR-10 ($r=5\%$).}
\end{subfigure}
\caption{Accuracy versus non-IID degree.}
\label{nonIID}
\end{figure}

Figure \ref{nonIID} illustrates model performance on the Fashion-MNIST and CIFAR-10 datasets across various non-IID degrees (with smaller $\alpha$ indicating higher non-IID degree). The results show that SemiDFL consistently outperforms other baselines (except DFL-UB) with higher accuracy under different non-IID degrees. Although almost all methods experience performance declines as the non-IID degree increases, SemiDFL exhibits a smaller decrease in accuracy in highly non-IID scenarios on easy tasks (e.g. Fashion-MNIST dataset), demonstrating its robustness over non-IID data distribution among clients.

\subsubsection{Robustness on various labeled data ratios}
\begin{figure}[t!]
\centering
\begin{subfigure}{0.48\linewidth}
	\centering
	\includegraphics[width=1\textwidth]{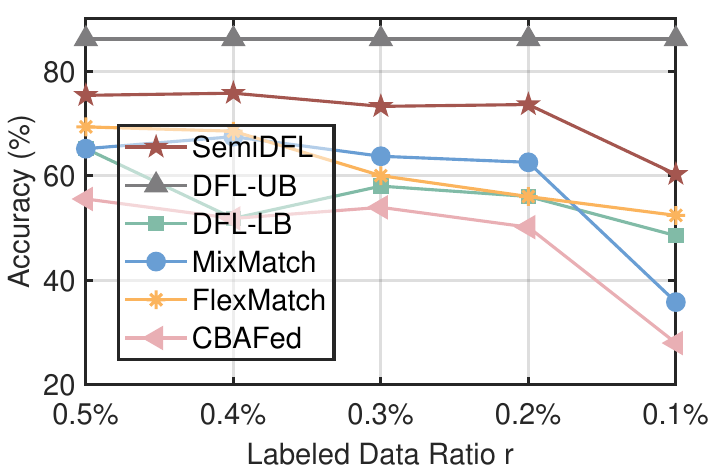}
	\caption{Fashion-MNIST.}
\end{subfigure}
\begin{subfigure}{0.48\linewidth}
	\centering
	\includegraphics[width=1\textwidth]{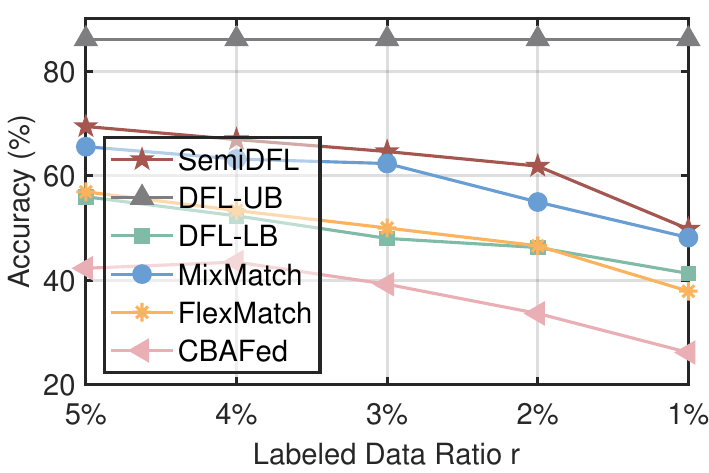}
	\caption{CIFAR-10.}
\end{subfigure}
\caption{Accuracy versus labeled data ratio ($\alpha=100$).}
\label{label_ratio}
\end{figure}

Figure \ref{label_ratio} demonstrates the accuracy of different models across various labeled data ratios on two datasets. Our method consistently outperforms other baselines (except DFL-UB) and exhibits a smaller performance decrease with a low labeled data ratio, exhibiting its robustness on various labeled data ratios.

\subsection{Ablation Study}
\subsubsection{Neighborhood Pseudo-Labeling} 

To further verify the effectiveness of neighborhood pseudo-labeling, we compare the following three variants of pseudo-labeling methods.
\begin{itemize}
    \item \textbf{Vanilla Pseudo-Labeling (Vanilla PL)}: each client takes the vanilla pseudo-labeling method without neighborhood classifier and filtering threshold.
    \item \textbf{Adaptive Pseudo-Labeling (APL)}: each client operates a pseudo-labeling method without neighborhood information, while the filtering threshold is adaptively updated based on the local number of qualified pseudo-labeled data.
    \item \textbf{Neighborhood Pseudo-Labeling (NPL, Ours)}: each client introduces neighborhood classifiers to enhance the pseudo-labeling process and uses neighborhood-qualified pseudo-label numbers to adaptively update the filtering threshold. 
\end{itemize}

Table \ref{ab:npl} shows that NPL consistently outperforms both Vanilla PL and Adaptive APL across all settings. 
While APL  improves Vanilla PL by incorporating an adaptive filtering threshold, it still lags in performance within the DFL scenario due to dynamic and noisy supervision, especially in highly non-IID cases. NPL significantly enhances model performance by leveraging neighborhood information and the number of qualified pseudo-labels in each round to refine supervision. This verifies the effectiveness and superiority of the proposed neighborhood pseudo-labeling method.

\begin{table}[t]
\centering
\footnotesize
\setlength{\tabcolsep}{2pt}
\caption{\footnotesize Ablation of neighborhood pseudo-labeling on CIFAR-10.}
\label{ab:npl}
\resizebox{0.95\linewidth}{!}{
\begin{tabular}{>{\centering\arraybackslash}m{0.115\textwidth-2\tabcolsep}>{\centering\arraybackslash}m{0.115\textwidth-2\tabcolsep}>{\centering\arraybackslash}m{0.115\textwidth-2\tabcolsep}>{\centering\arraybackslash}m{0.115\textwidth-2\tabcolsep}}
\toprule
\diagbox[dir=SE]{\scriptsize Setting}{\scriptsize Method} & Vanilla PL & APL & NPL (Ours) \\ 
\toprule
$\alpha=100$ $r=5\%$
& $40.51\pm13.74$ & $47.40\pm9.83$ & \cellcolor{gray!20} $\mathbf{69.40\pm0.20}$ \\ 
\midrule
$\alpha=100$ $r=1\%$
& $32.13\pm0.67$ & $34.88\pm0.89$ & \cellcolor{gray!20} $\mathbf{49.79\pm0.18}$ \\
\midrule
$\alpha=0.1$ $r=5\%$
& $24.51\pm3.98$ & $31.14\pm5.36$ & \cellcolor{gray!20} $\mathbf{49.63\pm0.61}$ \\
\midrule
$\alpha=0.1$ $r=1\%$
& $14.95\pm2.22$ & $30.89\pm1.03$ & \cellcolor{gray!20} $\mathbf{40.28\pm0.89}$ \\
\bottomrule
\end{tabular}
}
\vspace{-1em}
\end{table}

\subsubsection{Consensus MixUp}
To demonstrate the efficacy of the proposed C-MixUp method, we compare the following three variants. Notably, a client's labeled and pseudo-labeled data can be an empty set if it lacks this data source.

\begin{itemize}
\item \textbf{Local MixUp (L-MixUp)}: Each client performs local MixUp using labeled and pseudo-labeled data to generate mixed data for local classifier training.

\item \textbf{C-MixUp with GAN (w/ GAN)}: Each client performs C-MixUp leveraging labeled data, pseudo-labeled data, and synthesized data generated by a consensus-based GAN model \cite{odena2017conditional} to create mixed data for local classifier training.

\item \textbf{C-MixUp with Diffusion (w/ Diffusion, Ours)}: Each client performs C-MixUp using labeled data, pseudo-labeled data, and synthesized data generated by a consensus-based diffusion model to produce mixed data for local classifier training.
\end{itemize}

Table \ref{ab:mixup} shows that our proposed C-MixUp method consistently outperforms all three variants. Additionally, both C-MixUp with diffusion and GAN achieve higher accuracy than L-MixUp, particularly in highly non-iid scenarios (small $\alpha$). This verifies the efficacy of the consensus data space designed by C-MixUp in semi-supervised DFL tasks.

\begin{table}[t]
\centering
\footnotesize
\setlength{\tabcolsep}{2pt}
\caption{\footnotesize Ablation of Consensus-MixUp on MNIST.}
\label{ab:mixup}
\resizebox{0.95\linewidth}{!}{
\begin{tabular}{>{\centering\arraybackslash}m{0.12\textwidth-2\tabcolsep}>{\centering\arraybackslash}m{0.115\textwidth-2\tabcolsep}>{\centering\arraybackslash}m{0.115\textwidth-2\tabcolsep}>{\centering\arraybackslash}m{0.115\textwidth-2\tabcolsep}}
\toprule
\diagbox[dir=SE]{\scriptsize Setting}{\scriptsize Method} & L-MixUp & w/ GAN & w/ Diffusion (Ours)\\ 
\toprule
$\alpha=100$ $r=0.5\%$ 
& $93.69\pm0.09$ & $92.17\pm0.20$ & \cellcolor{gray!20} $\mathbf{94.46\pm0.06}$ \\ 
\midrule
$\alpha=100$ $r=0.1\%$ 
& $76.30\pm0.54$ & $90.29\pm0.10$ & \cellcolor{gray!20} $\mathbf{90.88\pm0.07}$ \\ 
\midrule
$\alpha=0.1$ $r=0.5\%$ 
& $70.75\pm9.39$ & $86.80\pm0.47$ & \cellcolor{gray!20} $\mathbf{88.69\pm0.09}$ \\ 
\midrule
$\alpha=0.1$ $r=0.1\%$ 
& $49.13\pm4.93$ & $69.32\pm0.89$ & \cellcolor{gray!20} $\mathbf{70.66\pm0.50}$ \\
\bottomrule
\end{tabular}
}
\end{table}

\subsubsection{Adaptive Aggregation}
We examine the efficacy of adaptive aggregation by comparison with its three variants:

\begin{itemize}
\item \textbf{Constant weight}: Clients in a sub-graph use a constant weight to aggregate their classifier and diffusion models. 
\item \textbf{Adaptive on test dataset (AdaTest)}: Clients aggregate models based on adaptive weights determined by their accuracy on an extra test dataset. For this ablation study only, we assume clients have access to an extra shared test dataset.
\item \textbf{Adaptive on generated dataset (AdaGen, Ours)}: All clients aggregate models based on adaptive weights determined by their accuracy on the dataset generated by diffusion models.
\end{itemize}

The results in Table \ref{ab:aa} demonstrate that the proposed AdaGen outperforms constant weight aggregation and shows comparable performance to AdaTest, despite not having access to an extra shared test dataset.
This enables effective performance evaluation while preserving data privacy.

\begin{table}[t]
\centering
\footnotesize
\setlength{\tabcolsep}{2pt}
\caption{\footnotesize {Ablation of adaptive aggregation on Fashion-MNIST}.}
\label{ab:aa}
\resizebox{0.95\linewidth}{!}{
\begin{tabular}{>{\centering\arraybackslash}m{0.115\textwidth-2\tabcolsep}>{\centering\arraybackslash}m{0.115\textwidth-2\tabcolsep}>{\centering\arraybackslash}m{0.115\textwidth-2\tabcolsep}>{\centering\arraybackslash}m{0.115\textwidth-2\tabcolsep}}
\toprule
\diagbox[dir=SE]{\scriptsize Setting}{\scriptsize Method} & Constant & AdaTest & AdaGen (Ours) \\ 
\toprule
$\alpha=100$ $r=0.5\%$
& $75.18\pm0.22$ & $\mathbf{75.80\pm0.12}$ & \cellcolor{gray!20} $75.36\pm0.27$ \\ 
\midrule
$\alpha=100$ $r=0.1\%$
& $56.12\pm0.11$ & $59.03\pm0.13$ & \cellcolor{gray!20} $\mathbf{60.27\pm0.16}$ \\ 
\midrule
$\alpha=0.1$ $r=0.5\%$
& $70.06\pm0.34$ & $70.46\pm0.78$  & \cellcolor{gray!20} $\mathbf{71.49\pm0.14}$ \\ 
\midrule
$\alpha=0.1$ $r=1\%$
& $49.11\pm0.56$ & $\mathbf{51.38\pm0.64}$ & \cellcolor{gray!20} $49.40\pm0.71$ \\
\bottomrule
\end{tabular}
}
\vspace{-1em}
\end{table}

\section{Conclusion}
\label{s7}
This paper proposes SemiDFL, the first semi-supervised DFL paradigm through constructing consensus data and model spaces among clients to tackle the challenges of limited labels and highly non-IID data distributions in DFL. 
SemiDFL utilizes neighborhood information to enhance the estimation and filtering of pseudo-labels for unlabeled samples, improving both the quality and robustness of pseudo-labeling. 
Additionally, a consensus-based diffusion model generates synthesized data with a similar distribution, facilitating MixUp and forming a consensus data space that mitigates non-IID issues in classifier training. 
We further design an adaptive aggregation strategy based on each client's performance to establish a more effective consensus model space, enhancing classifier performance. 
Extensive experimental evaluations demonstrate the superior performance of SemiDFL compared to existing semi-supervised learning methods in DFL scenarios.

\section{Acknowledgements}
\label{s8}
This work is supported by National Natural Science Foundation of China (Grant No.62203309, 62401161), Guangdong Basic and Applied Basic Research Foundation (Grant No. 2024A1515011333, 2022A1515110056), Shenzhen Science and Technology Program (Grant No. RCBS20221008093312031), Longgang District Shenzhen's “Ten Action Plan”for Supporting Innovation Projects (Grant No. LGKCSDPT2024002, LGKCSDPT2024003), and the Shenzhen Institute of Artificial Intelligence and Robotics for Society.

\bibliography{aaai25}
\externaldocument{Supplementary}
\end{document}


\date{\vspace{-6ex}}
\maketitle

This supplementary document provides detailed experimental settings, results and analyses, which have not been included in the main paper due to the page limit. 
Specifically, we first present more details of the experimental setup and related hyper-parameters.
Then, we present more evaluation results under different communication topologies. 
Our source code will be made publicly available upon acceptance of the paper.

\section{Experimental Setup}
Taking Figure~1 (a) in the main paper as an example of the decentralized communication topology, we evaluate SemiDFL on different datasets, various labeled data ratios, and non-IID degrees.
Specifically, we evaluate SemiDFL on MNIST~\cite{lecun1998mnist} and Fashion-MNIST~\cite{xiao2017fashion} using Convolutional Neural Network (CNN), and on CIFAR-10~\cite{krizhevsky2010cifar} using ResNet-18. 
We adopt the same method in \cite{hsu2019measuring,lin2020ensemble} to simulate different non-IID data distribution degrees. 
The non-IID degree is captured by a sample allocation probability $\alpha$, with smaller $\alpha$ indicating a higher non-IID degree. 
The percentage of total labeled data in the union of all clients' data is denoted by the labeled data ratio $r$.
We train models for $500$ global rounds on all datasets.  All experiments are conducted using PyTorch $2.0$ on a machine with $2$ RTX $4090$ GPUs. The detailed experimental settings and parameters are as follows.

\subsection{Datasets} 
We evaluate our proposed SemiDFL on the following three benchmark datasets. 
\begin{itemize}
\item \textbf{MNIST}: 
A classic dataset for handwritten digit classification, comprising 70,000 grayscale images (60,000 for training and 10,000 for testing) across 10 classes (digits 0-9). 
It’s widely used to assess basic image classification models \cite{lecun1998mnist}. 
\item \textbf{Fashion-MNIST}:
Serving as a more complex alternative to MNIST, this dataset consists of 70,000 images, split into 60,000 for training and 10,000 for testing. 
It features 10 classes of fashion items, such as clothing and accessories, offering a slightly more challenging classification task \cite{xiao2017fashion}.
\item \textbf{CIFAR-10}:
This dataset includes 60,000 images across 10 classes (e.g., animals, vehicles), with 5,000 images per class for training and 1,000 for testing. 
CIFAR-10 is used to evaluate models on more diverse and complex image classification tasks \cite{krizhevsky2010cifar}.
\end{itemize}

\begin{figure}[h!]
\centering
\begin{subfigure}{1\textwidth}
	\centering
	\includegraphics[width=\textwidth]{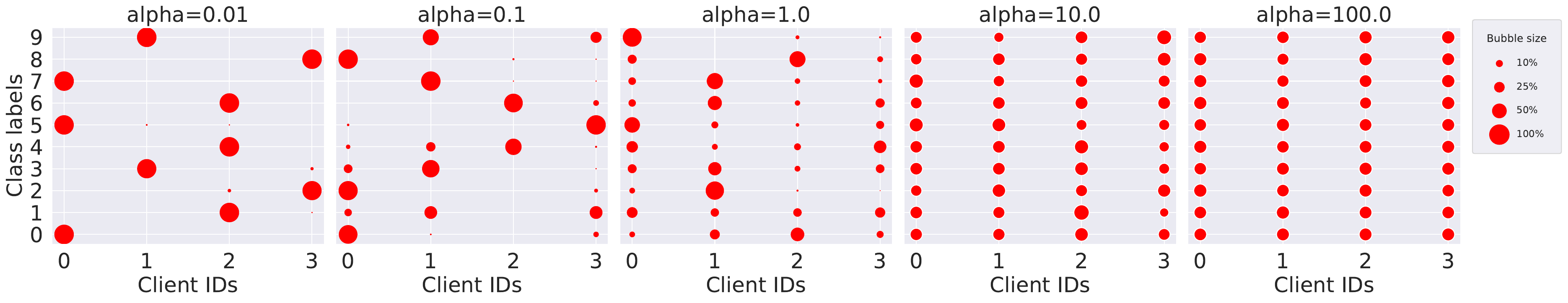}
	\caption{Class distribution of labeled data over L- and M- clients on CIFAR10 dataset when $r=5$.}
\end{subfigure}
\begin{subfigure}{1\textwidth}
	\centering
	\includegraphics[width=\textwidth]{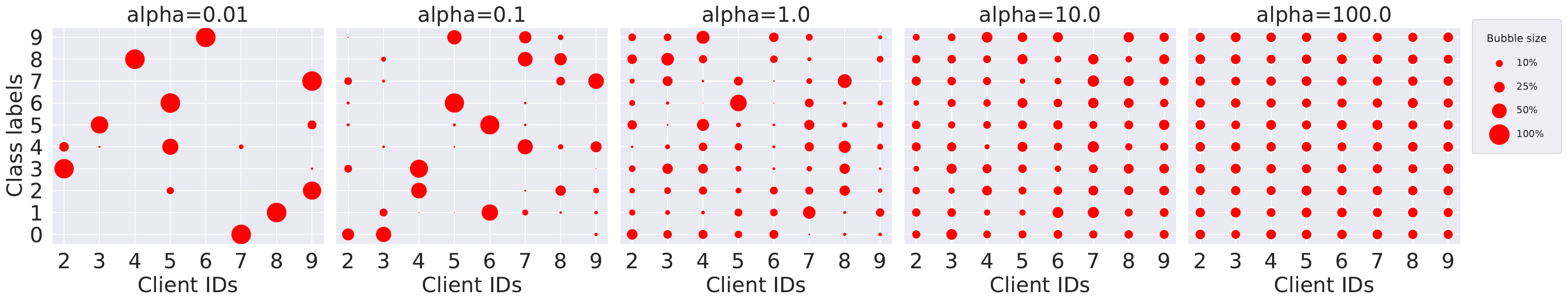}
	\caption{Class distribution of unlabeled data over M- and U- clients on CIFAR10 dataset when $r=5$.}
\end{subfigure}
\begin{subfigure}{1\textwidth}
	\centering
	\includegraphics[width=\textwidth]{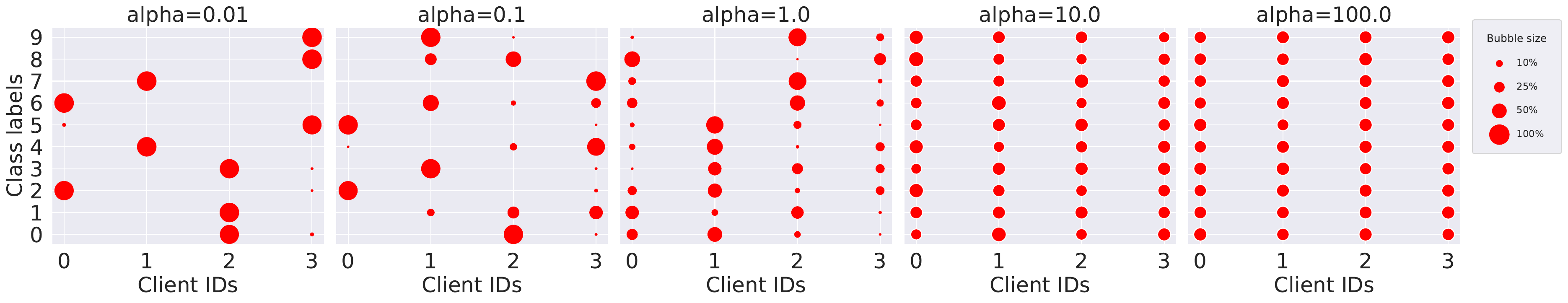}
	\caption{Class distribution of labeled data over L- and M- clients on CIFAR10 dataset when $r=1$.}
\end{subfigure}
\begin{subfigure}{1\textwidth}
	\centering
	\includegraphics[width=\textwidth]{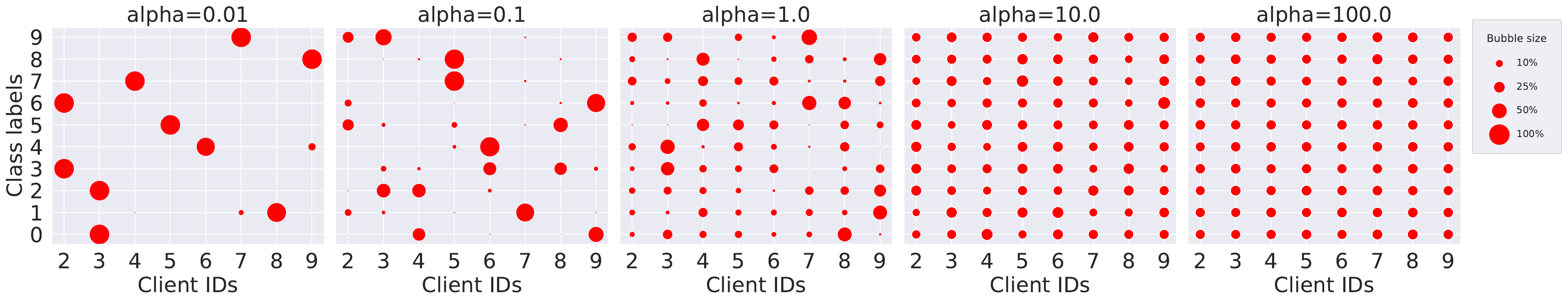}
	\caption{Class distribution of unlabeled data over M- and U- clients on CIFAR10 dataset when $r=1$.}
\end{subfigure}

\caption{The non-IID data distribution of CIFAR10 dataset under different settings.}
\label{non_iid_distribution}
\end{figure}

Take the CIFAR10 dataset as an example, we visualize the non-IID partition and its data distribution under the topology of primary experiments in the main paper, as shown in the Figure \ref{non_iid_distribution}.

\subsection{Models}
For MNIST and Fashion-MNIST, we train a CNN (Conv2d($1*6*3$) → ReLU → MaxPool2d($2*2$) → Conv2d($6*25*3$) → ReLU → MaxPool2d($2*2$) → Linear($1225*50$) → ReLU → Linear($50*10$)), as for CIFAR-10, we use ResNet-18. 
For the implementation of DDPM, we use the UNet to predict the noise as proposed in \cite{ho2020denoising}.
To facilitate the sampling speed and decrease the computation cost of data generation, we utilize the ODE-solver proposed in \cite{lu2022dpm} to reduce the number of sampling steps required by DDPM.
For the implementation of GAN model in the ablation study of main paper, we use two CNNs as the discriminator and generator respectively \cite{odena2017conditional}.
Notably, we replace all of the BatchNorm with GroupNorm to avoid the training issue caused by BatchNorm \cite{zhang2021flexmatch}.

\subsection{Hyper-Parameters}
In each global training round, each client performs $E$ ($25$ for MNIST, $50$ for Fashion-MNIST and CIFAR-$10$) iterations of local training via mini-batch SGD with a batch size of $B = 10$. 
Other hyper-parameters of optimizer during local model training are inherited from the default settings of Adam~\cite{kingma2014adam} and for all datasets, we use a learning rate of 0.05 to train classifiers, and 
the learning rate of DDPM training is set to 0.001.
The sharpen temperature $Z$ and the qualification threshold $\tau$ of pseudo labeling is set to $2.0$ and $0.95$ respectively.
For MixUp operation, the $\lambda$ is sampled from the distribution $\text{Beta}(0.5, 0.5)$ \cite{berthelot2019mixmatch}. To avoid the computation overhead, we generate $1000$ samples per 10 global rounds for each client to train the classifier, and $100$ samples to evaluate the performance of classifier.

\section{Robustness on different decentralized communication topologies}
To comprehensively assess the influence of communication topology, we conduct evaluations under different topologies and use the same SSL methods from the main paper as baselines.
We employ Topologies 1-3 for the MNIST, Fashion-MNIST, and CIFAR-10 datasets, respectively.
The description of Topologies 1-3 are as follows, with their diagram shown in Figure \ref{non_iid_distribution}.
\begin{itemize}
    \item \textbf{Topology 1} We change the number of clients for different types, with only one labeled client, six unlabeled client, and three mixed client.

    \item \textbf{Topology 2} We consider a special case where two labeled clients are connected, each linked to three unlabeled clients and one mixed client.

    \item \textbf{Topology 3} We consider a ring topology with three labeled clients, four unlabeled clients and three mixed clients. Each client connects to two neighbors, with labeled clients separated by unlabeled and mixed clients.
\end{itemize}

\begin{figure}[ht!]
\centering
\begin{subfigure}{0.31\textwidth}
	\centering
	\includegraphics[width=\textwidth]{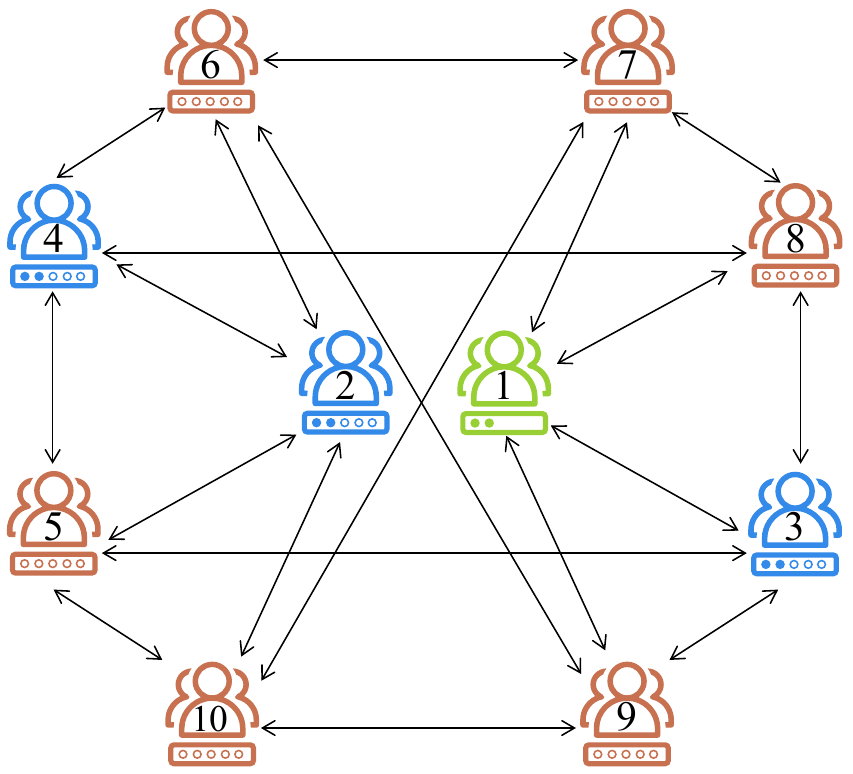}
	\caption{Topology 1.}
\end{subfigure}
\hspace{0.025\textwidth}
\begin{subfigure}{0.31\textwidth}
	\centering
	\includegraphics[width=\textwidth]{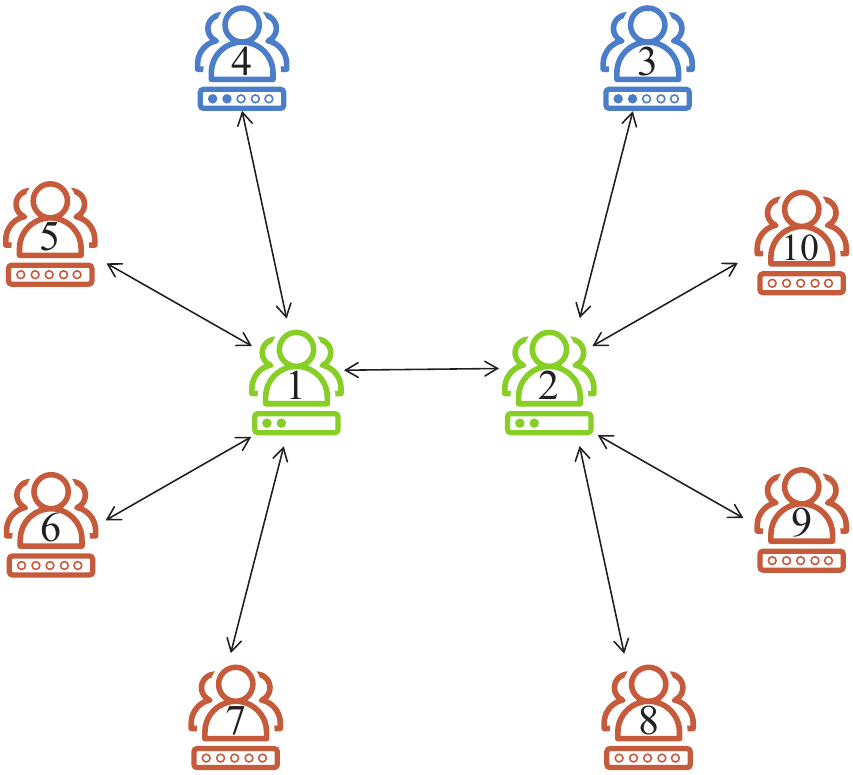}
	\caption{Topology 2.}
\end{subfigure}
\hspace{0.025\textwidth}
\begin{subfigure}{0.31\textwidth}
	\centering
	\includegraphics[width=\textwidth]{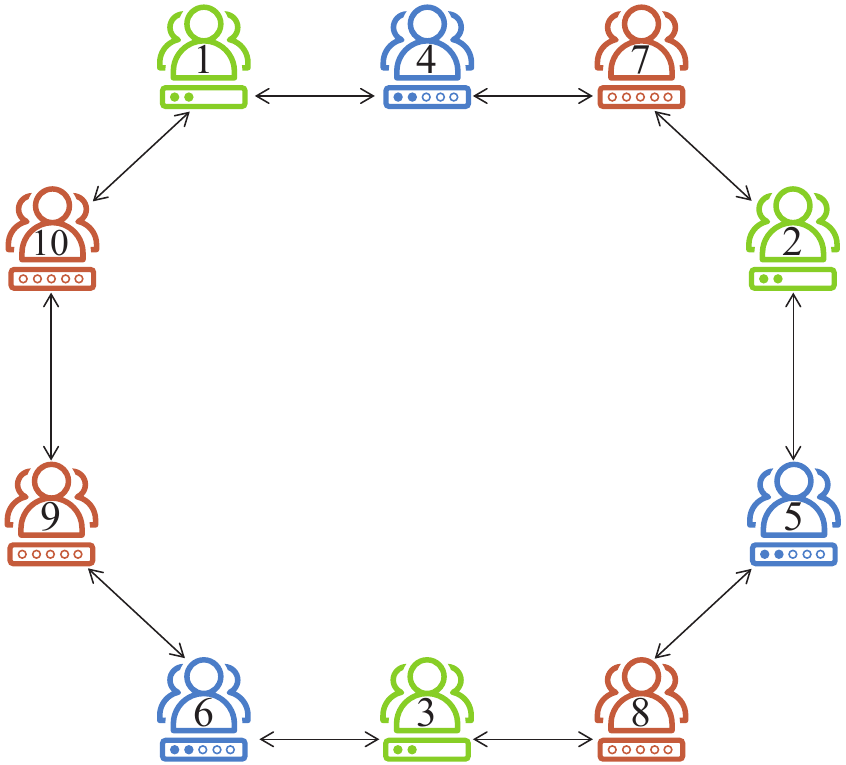}
	\caption{Topology 3.}
\end{subfigure}
\caption{Evaluated communication topologies.}
\label{topologies}
\end{figure}

Table \ref{topology_exp} presents the quantitative results of our SemiDFL alongside the state-of-the-art baselines over topologies 1-3 on three datasets. 
It is seen from the table that our proposed SemiDFL still consistently outperforms all baselines (except DFL-UB) across different communication topologies with various non-IID degrees $\alpha$ and labeled data ratios $r$.
This further verifies the robustness of the proposed SemiDFL on different decentralized communication topologies.

\begin{table*}[ht!]
\centering
\footnotesize
\caption{Average accuracy $\pm$ Standard deviation of all clients across MNIST, Fashion-MNIST, and CIFAR-10 datasets, evaluated under different communication topologies. (DFL-UB assumes all data are labeled, so it remains unchanged with varying $r$. The symbol ``-----"  denotes a non-convergent result.)}
 
\label{topology_exp}
\begin{tabular}{cccccccc}

\toprule
\multirow{2}{*}{\diagbox[dir=SE]{\footnotesize Method}{\footnotesize Setting}} & \multirow{2}{0.1\textwidth-2\tabcolsep - 1.25\arrayrulewidth}{\centering non-IID Degree} & \multicolumn{2}{c}{MNIST / Topology 1} & \multicolumn{2}{c}{Fashion-MNIST / Topology 2} & \multicolumn{2}{c}{CIFAR-10 / Topology 3} \\ 
\cmidrule(l){3-4} \cmidrule(l){5-6} \cmidrule(l){7-8}
& & $r=0.5\%$ &  $r=0.1\%$ & $r=0.5\%$ & $r=0.1\%$ & $r=5\%$ & $r=1\%$ \\ \toprule
\multirow{2}{*}{DFL-UB} 
& $\alpha=100$ & \multicolumn{2}{c}{$95.76\pm0.10$}& \multicolumn{2}{c}{$83.68\pm0.71$} & \multicolumn{2}{c}{$79.77\pm0.14$}\\ 
& $\alpha=0.1$ & \multicolumn{2}{c}{$94.27\pm0.76$}& \multicolumn{2}{c}{$59.49\pm4.91$} & \multicolumn{2}{c}{$44.60\pm2.19$}\\ 
\midrule
\multirow{2}{*}{DFL-LB} 
& $\alpha=100$ & $92.29\pm0.18$ & $69.98\pm0.01$ & $73.00\pm0.57$ & $64.80\pm0.23$ & $53.65\pm0.17$ & $39.98\pm0.13$ \\ 
& $\alpha=0.1$ & $84.96\pm0.90$ & $67.51\pm0.02$ & $58.37\pm5.73$ & $52.12\pm2.14$ & $34.15\pm2.03$ & $31.51\pm1.83$ \\
\midrule
\multirow{2}{*}{MixMatch} 
& $\alpha=100$ & $88.73\pm1.25$ & $32.19\pm10.53$ & $66.47\pm6.53$ & $35.84\pm9.88$ & $63.87\pm0.45$ & $46.40\pm0.63$\\ 
& $\alpha=0.1$ & $72.25\pm5.12$ & $28.76\pm9.22$ & $36.48\pm7.43$ & $25.31\pm7.51$ & $38.21\pm5.28$ & $26.23\pm2.39$\\
\midrule
\multirow{2}{*}{FlexMatch} 
& $\alpha=100$ & $94.67\pm0.05$ & $64.04\pm0.90$ & $71.63\pm1.94$ & $47.38\pm7.20$ & $55.11\pm0.34$ & $38.17\pm0.56$\\ 
& $\alpha=0.1$ & $83.92\pm3.08$ & $53.88\pm3.97$ & $37.57\pm10.07$ & $27.37\pm8.60$ & $39.49\pm2.93$ & $28.50\pm0.70$\\
\midrule
\multirow{2}{*}{CBAFed} 
& $\alpha=100$ & $71.60\pm3.14$ & $38.87\pm0.13$ & $69.37\pm1.12$ & $41.06\pm3.19$ & $44.29\pm3.42$ & $26.05\pm1.44$\\ 
& $\alpha=0.1$ & $61.06\pm3.72$ & $30.77\pm1.64$ & $44.47\pm4.09$ & $33.35\pm3.24$ & $33.04\pm3.20$ & $24.87\pm2.09$\\
\midrule
\rowcolor{gray!20} 
& $\alpha=100$ & $\mathbf{95.06\pm0.05}$ & $\mathbf{79.32\pm0.06}$ & $\mathbf{77.40\pm0.92}$ & $\mathbf{68.20\pm1.48}$ & $\mathbf{70.30\pm0.45}$ & $\mathbf{52.54\pm0.51}$ \\
\rowcolor{gray!20} \multirow{-2}{*}{\textbf{SemiDFL}} 
& $\alpha=0.1$ & $\mathbf{88.93\pm0.16}$ & $\mathbf{70.84\pm0.51}$ & $\mathbf{68.21\pm1.37}$ & $\mathbf{63.39\pm1.73}$ & $\mathbf{53.99\pm2.11}$ & $\mathbf{39.96\pm1.67}$ \\ 
\bottomrule
\end{tabular}
\end{table*}

\section{Reproducibility Checklist}

This paper:

\begin{itemize}
    \item Includes a conceptual outline and/or pseudocode description of AI methods introduced. (yes)
    \item Clearly delineates statements that are opinions, hypothesis, and speculation from objective facts and results. (yes)
    \item Provides well marked pedagogical references for less-familiare readers to gain background necessary to replicate the paper. (yes)
\end{itemize}

Does this paper make theoretical contributions? (no)

Does this paper rely on one or more datasets? (yes)

\begin{itemize}
    \item A motivation is given for why the experiments are conducted on the selected datasets. (yes)
    \item All novel datasets introduced in this paper are included in a data appendix. (NA)
    \item All novel datasets introduced in this paper will be made publicly available upon publication of the paper with a license that allows free usage for research purposes. (NA)
    \item All datasets drawn from the existing literature (potentially including authors’ own previously published work) are accompanied by appropriate citations. (yes)
    \item All datasets drawn from the existing literature (potentially including authors’ own previously published work) are publicly available. (yes)
    \item All datasets that are not publicly available are described in detail, with explanation why publicly available alternatives are not scientifically satisficing. (NA)
\end{itemize}

Does this paper include computational experiments? (yes)

\begin{itemize}
    \item  Any code required for pre-processing data is included in the appendix. (no).
    \item  All source code required for conducting and analyzing the experiments is included in a code appendix. (no)
    \item All source code required for conducting and analyzing the experiments will be made publicly available upon publication of the paper with a license that allows free usage for research purposes. (yes)
    \item All source code implementing new methods have comments detailing the implementation, with references to the paper where each step comes from. (yes)
    \item If an algorithm depends on randomness, then the method used for setting seeds is described in a way sufficient to allow replication of results. (yes)
    \item This paper specifies the computing infrastructure used for running experiments (hardware and software), including GPU/CPU models; amount of memory; operating system; names and versions of relevant software libraries and frameworks. (yes)
    \item This paper formally describes evaluation metrics used and explains the motivation for choosing these metrics. (yes)
    \item This paper states the number of algorithm runs used to compute each reported result. (yes)
    \item Analysis of experiments goes beyond single-dimensional summaries of performance (e.g., average; median) to include measures of variation, confidence, or other distributional information. (yes)
    \item The significance of any improvement or decrease in performance is judged using appropriate statistical tests (e.g., Wilcoxon signed-rank). (yes)
    \item This paper lists all final (hyper-)parameters used for each model/algorithm in the paper’s experiments. (yes)
    \item This paper states the number and range of values tried per (hyper-) parameter during development of the paper, along with the criterion used for selecting the final parameter setting. (partial)

\end{itemize}

\newpage
\bibliographystyle{IEEEtran}
\bibliography{aaai25}